\pdfoutput=1
\documentclass[11pt]{article}

\usepackage{acl}

\usepackage{times}
\usepackage{latexsym}

\usepackage[T1]{fontenc}
\usepackage[utf8]{inputenc}

\usepackage{microtype}

\usepackage{hyperref}
\usepackage{url}
\usepackage{booktabs}
\usepackage{tablefootnote}
\usepackage{multirow}
\usepackage{makecell}
\usepackage{graphicx}
\usepackage{subcaption}
\usepackage{amsmath}
\usepackage{arydshln}
\usepackage{amssymb}

\title{MarkupLM: Pre-training of Text and Markup Language \\for Visually Rich Document Understanding}

\author{Junlong Li$^{1}$\thanks{\ \ Equal contributions during internship at Microsoft Research Asia. Corresponding authors: Lei Cui and Furu Wei}\ , Yiheng Xu$^{2*}$,  Lei Cui$^{2}$, Furu Wei$^{2}$ \\ 
$^{1}$Shanghai Jiao Tong University\\
$^{2}$Microsoft Research Asia\\
\texttt{lockonn@sjtu.edu.cn}\\
\texttt{\{t-yihengxu,lecu,fuwei\}@microsoft.com} \\
}

\begin{document}
\maketitle
\begin{abstract}
Multimodal pre-training with text, layout, and image has made significant progress for Visually Rich Document Understanding (VRDU), especially the fixed-layout documents such as scanned document images. While, there are still a large number of digital documents where the layout information is not fixed and needs to be interactively and dynamically rendered for visualization, making existing layout-based pre-training approaches not easy to apply. In this paper, we propose \textbf{MarkupLM} for document understanding tasks with markup languages as the backbone, such as HTML/XML-based documents, where text and markup information is jointly pre-trained. Experiment results show that the pre-trained MarkupLM significantly outperforms the existing strong baseline models on several document understanding tasks. The pre-trained model and code will be publicly available at \url{https://aka.ms/markuplm}.
\end{abstract}

\section{Introduction}

Multimodal pre-training with text, layout, and visual information has recently become the de facto approach~\citep{10.1145/3394486.3403172,xu2021layoutlmv2,xu2021layoutxlm,pramanik2020multimodal,garncarek2021lambert,hong2021bros,powalski2021going,wu2021lampret,li2021structurallm,li2021selfdoc,appalaraju2021docformer} in Visually-rich Document Understanding (VRDU) tasks. These multimodal models are usually pre-trained with the Transformer architecture~\citep{vaswani2017attention} using large-scale unlabeled scanned document images~\citep{10.1145/1148170.1148307} or digital-born PDF files, followed by task-specific fine-tuning with relatively small-scale labeled training samples to achieve the state-of-the-art performance on a variety of document understanding tasks, including form understanding~\citep{Jaume2019FUNSDAD,xu2021layoutxlm}, receipt understanding~\citep{8977955,park2019cord}, complex document understanding~\citep{graliski2020kleister}, document type classification~\citep{harley2015icdar}, and document visual question answering~\citep{mathew2020docvqa}, etc. Significant progress has been witnessed not only in research tasks within academia, but also in different real-world business applications such as finance, insurance, and many others.

\begin{figure*}[t]
\centering
    \begin{subfigure}[b]{0.12\textwidth}
        \frame{\includegraphics[width=\textwidth]{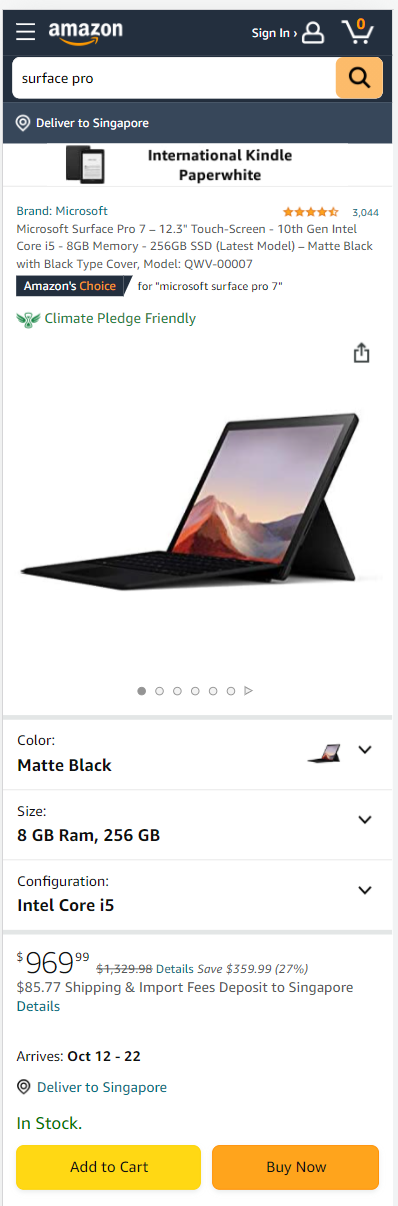}}
        \caption{Mobile}
        \label{fig:1a}
    \end{subfigure}
    ~%\qquad %add desired spacing between images, e. g. ~, \quad, \qquad, \hfill etc.
      %(or a blank line to force the subfigure onto a new line)
    \begin{subfigure}[b]{0.271\textwidth}
        \frame{\includegraphics[width=\textwidth]{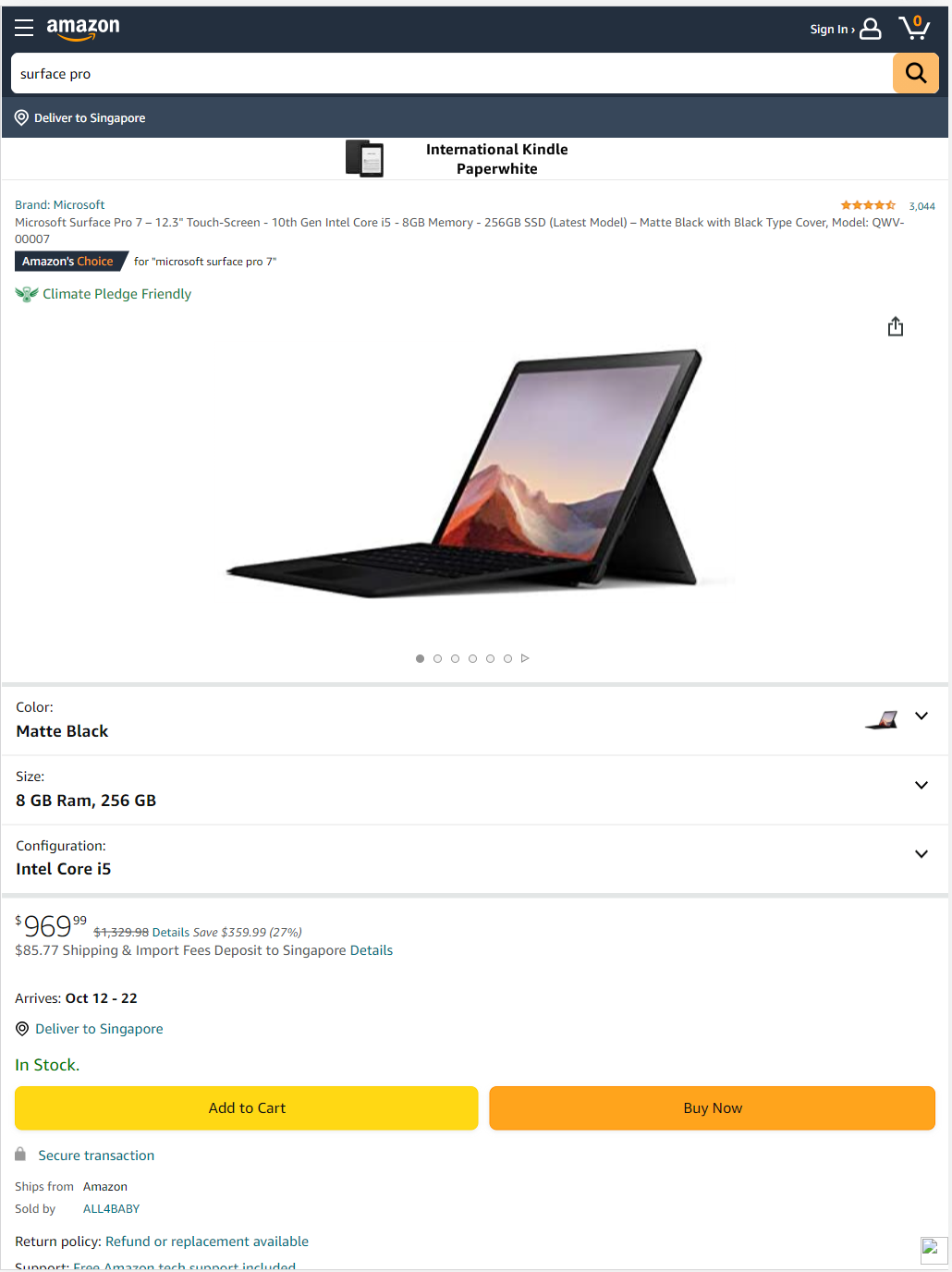}}
        \caption{Tablet}
        \label{fig:1b}
    \end{subfigure}
    ~%\qquad %add desired spacing between images, e. g. ~, \quad, \qquad, \hfill etc.
      %(or a blank line to force the subfigure onto a new line)
    \begin{subfigure}[b]{0.57\textwidth}
        \frame{\includegraphics[width=\textwidth]{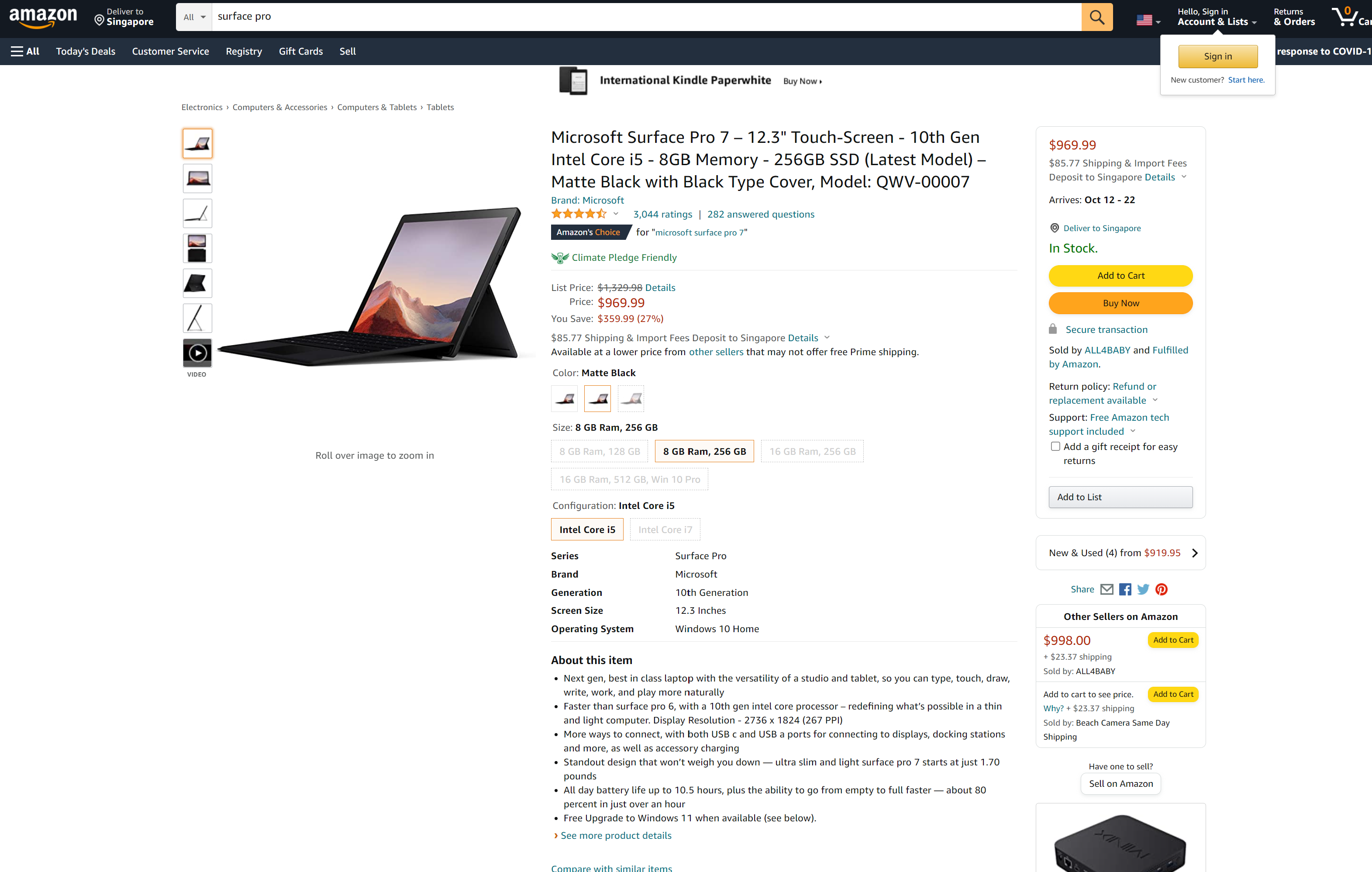}}
        \caption{Desktop}
        \label{fig:1c}
    \end{subfigure}
    ~ %add desired spacing between images, e. g. ~, \quad, \qquad, \hfill etc.
    %(or a blank line to force the subfigure onto a new line)
    % \begin{subfigure}[b]{0.35\textwidth}
    %     \frame{\includegraphics[width=\textwidth]{word.png}}
    %     \caption{Office Word}
    %     \label{fig:1c}
    % \end{subfigure}
    % ~\qquad
    % \begin{subfigure}[b]{0.29\textwidth}
    %     \frame{\includegraphics[width=\textwidth]{word_xml.png}}
    %     \caption{OOXML snippet}
    %     \label{fig:1d}
    % \end{subfigure}
    % \caption{HTML-based webpages rendered by different platforms, such as mobile, tablet and desktop. (\url{https://www.amazon.com/New-Microsoft-Surface-Pro-Touch-Screen/dp/B07YNHYQ5Z/ref=sr_1_3?dchild=1&keywords=surface+pro&qid=1633689243&sr=8-3})}
    \caption{HTML-based webpages rendered by different platforms, such as mobile, tablet and desktop. (\url{https://amzn.to/2ZZoi5R})}
    \label{fig:1}
\end{figure*}

Visually rich documents can be generally divided into two categories. The first one is the fixed-layout documents such as scanned document images and digital-born PDF files, where the layout and style information is pre-rendered and independent of software, hardware, or operating system. This property makes existing layout-based pre-training approaches easily applicable to document understanding tasks. While, the second category is the markup-language-based documents such as HTML/XML, where the layout and style information needs to be interactively and dynamically rendered for visualization depending on the software, hardware, or operating system, which is shown in Figure~\ref{fig:1}. For markup-language-based documents, the 2D layout information does not exist in an explicit format but usually needs to be dynamically rendered for different devices, e.g., mobile/tablet/desktop,
%In addition, the document style information is often explicitly described with the markup languages such as font type/size/color etc., 
which makes current layout-based pre-trained models hard to apply. Therefore, it is indispensable to leverage the markup structure into document-level pre-training for downstream VRDU tasks.

\begin{figure*}[t]
  \centering
  \includegraphics[width=\textwidth]{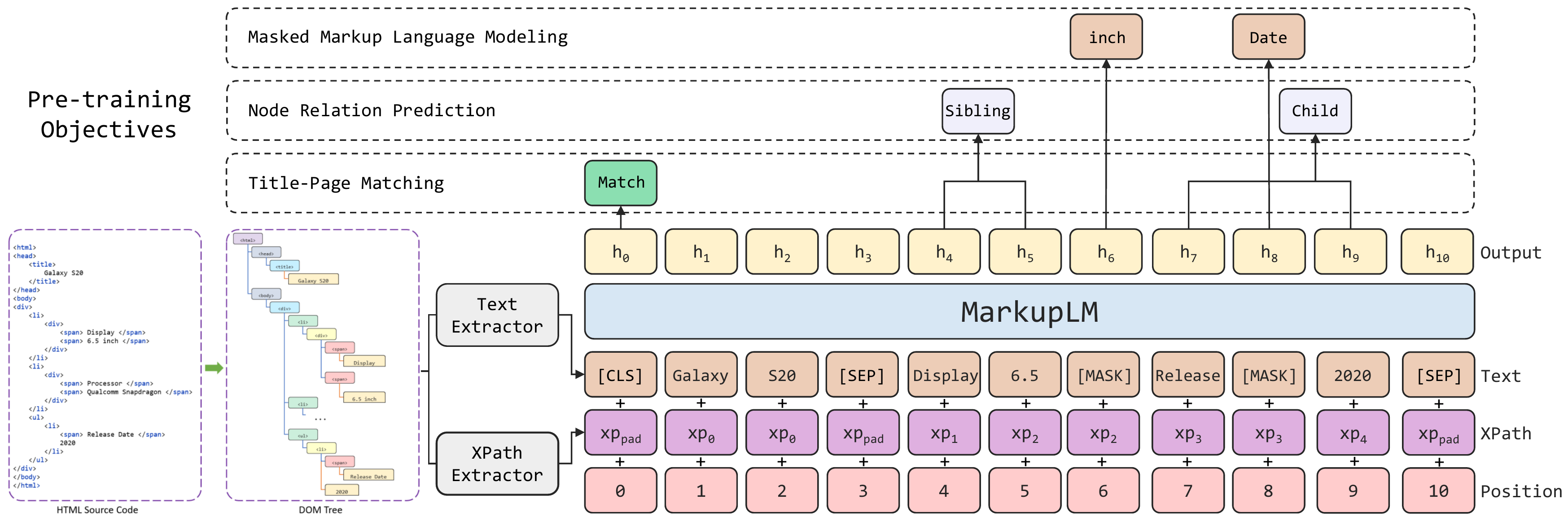}
 \caption{The architecture of MarkupLM, where the pre-training tasks are also included.}
 \label{fig:markuplm}
\end{figure*}

To this end, we propose \textbf{MarkupLM} to jointly pre-train text and markup language in a single framework for markup-based VRDU tasks.  Distinct from fixed-layout documents, markup-based documents provide another viewpoint for the document representation learning through markup structures because the 2D position information and document image information cannot be used straightforwardly during the pre-training. Instead, MarkupLM takes advantage of the tree-based markup structures to model the relationship among different units within the document. Similar to other multimodal pre-trained layout-based models, MarkupLM has four input embedding layers: (1) a text embedding that represents the token sequence information; (2) an XPath embedding that represents the markup tag sequence information from the root node to the current node; (3) a 1D position embedding that represents the sequence order information; (4) a segment embedding for downstream tasks. The overall architecture of MarkupLM is shown in Figure~\ref{fig:markuplm}. The XPath embedding layer can be considered as the replacement of 2D position embeddings compared with the LayoutLM model family ~\citep{10.1145/3394486.3403172,xu2021layoutlmv2,xu2021layoutxlm}. To effectively pre-train the MarkupLM, we use three pre-training strategies. The first is the Masked Markup Language Modeling (MMLM), which is used to jointly learn the contextual information of text and markups. The second is the Node Relationship Prediction (NRP), where the relationships are defined according to the hierarchy from the markup trees. The third is the Title-Page Matching (TPM), where the content within ``<title> ... </title>'' is randomly replaced by a title from another page to make the model learn whether they are correlated. In this way, MarkupLM can better understand the contextual information through both the language and markup hierarchy perspectives. 
%We pre-train two MarkupLM models with different structures, including an HTML-based pre-trained MarkupLM and an OOXML-based pre-trained MarkupLM. 
% Meanwhile, we select two benchmark datasets as the downstream tasks to evaluate the performance of MarkupLM, including the Structured Web Data Extraction
% (SWDE) dataset~\citep{10.1145/2009916.2010020} and the Web-based Structural Reading Comprehension (WebSRC) dataset~\citep{chen2021websrc}.
We evaluate the MarkupLM models on the Web-based Structural Reading Comprehension (WebSRC) dataset~\citep{chen-etal-2021-websrc} and the Structured Web Data Extraction (SWDE) dataset~\citep{10.1145/2009916.2010020}.
%an in-house page object detection dataset for Office Word documents as well as an in-house document type classification dataset for Office Word. 
Experiment results show that the pre-trained MarkupLM significantly outperforms the several strong baseline models in these tasks.

The contributions of this paper are summarized as follows:

\begin{itemize}
    \item We propose MarkupLM to address the document representation learning where the layout information is not fixed and needs to be dynamically rendered. For the first time, the text and markup information is pre-trained in a single framework for the VRDU tasks. 
    \item MarkupLM integrates new input embedding layers and pre-training strategies, which have been confirmed effective on HTML-based downstream tasks.
    \item The pre-trained MarkupLM models and code will be publicly available at \url{https://aka.ms/markuplm}.
\end{itemize}

% \begin{figure*}[t]
% \centering
%     \begin{subfigure}[b]{0.455\textwidth}
%         \frame{\includegraphics[width=\textwidth]{html2dom_example.pdf}}
%         \caption{DOM tree and XPath with the source HTML code}
%         \label{fig:2a}
%     \end{subfigure}
%     ~%\qquad %add desired spacing between images, e. g. ~, \quad, \qquad, \hfill etc.
%       %(or a blank line to force the subfigure onto a new line)
%     \begin{subfigure}[b]{0.53\textwidth}
%         \frame{\includegraphics[width=\textwidth]{xpath_emb.PNG}}
%         \caption{XPath embedding}
%         \label{fig:2b}
%     \end{subfigure}
%     ~%\qquad %add desired spacing between images, e. g. ~, \quad, \qquad, \hfill etc.
%       %(or a blank line to force the subfigure onto a new line)
%     \caption{An Example of the DOM tree and XPath, followed by the overview of the XPath embedding.}
%     \label{fig:2}
% \end{figure*}

\section{MarkupLM}
%MarkupLM aims to jointly pre-train the text and markup in markup-based visually rich documents and transfer the learned knowledge to downstream tasks.
MarkupLM utilizes the DOM tree in markup language and the XPath query language to obtain the markup streams along with natural texts in markup-language-based documents (Section \ref{section:dom}). We propose this Transformer-based model with a new XPath embedding layer to accept the markup sequence inputs (Section \ref{section:modelarch}) and pre-train it with three different-level objectives, including Masked Markup Language Modeling (MMLM), Node Relation Prediction (NRP), and Title-Page Matching (TPM) (Section \ref{section:pre-training}). 

\subsection{DOM Tree and XPath}
\label{section:dom}

% DOM Tree \footnote{\url{https://en.wikipedia.org/wiki/Document_Object_Model}}
% XPath \footnote{\url{https://en.wikipedia.org/wiki/XPath}}

A DOM\footnote{\url{https://en.wikipedia.org/wiki/Document_Object_Model}} tree is the tree structure object of a markup-language-based document (\textit{e.g.,} HTML or XML) in the view of DOM (Document Object Model) wherein each node is an object representing a part of the document. 

XPath\footnote{\url{https://en.wikipedia.org/wiki/XPath}} (XML Path Language) is a query language for selecting nodes from a markup-language-based document, which is based on the DOM tree and can be used to easily locate a node in the document. In a typical XPath expression, like \texttt{/html/body/div/li[1]/div/span[2]}, the texts stand for the tag name of the nodes while the subscripts are the ordinals of a node when multiple nodes have the same tag name under a common parent node.

\begin{figure}
    \centering
    \includegraphics[width=0.45\textwidth]{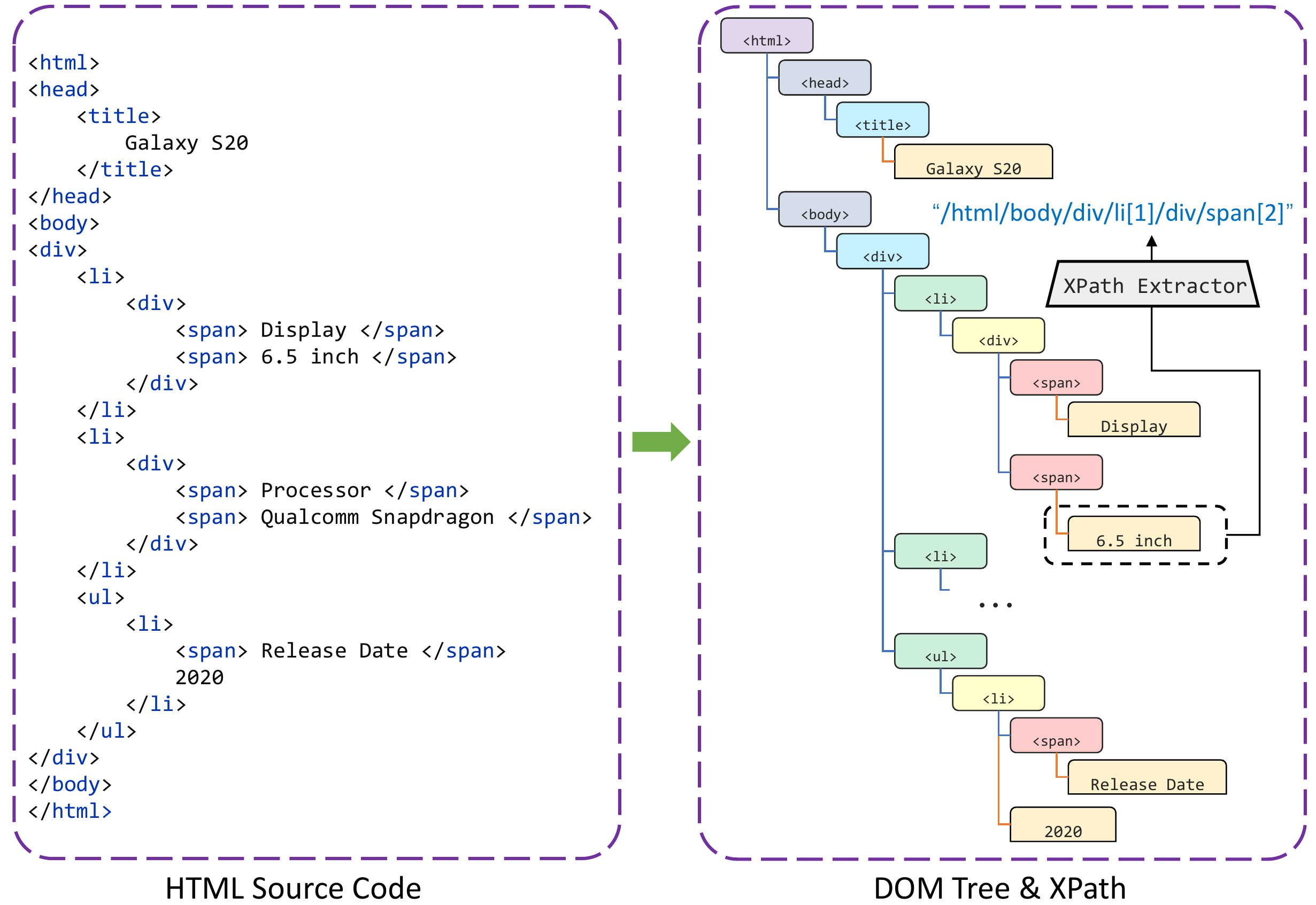}
    \caption{An example of DOM tree and XPath with the source HTML code. }
    \label{fig:dom_tree_example}
\end{figure}

We show an example of DOM tree and XPath along with the corresponding source code in Figure \ref{fig:dom_tree_example}, from which we can clearly identify the genealogy of all nodes within the document, as well as their XPath expressions.

\subsection{Model Architecture}
\label{section:modelarch}
\begin{figure*}
    \centering
    \includegraphics[width=0.85\textwidth]{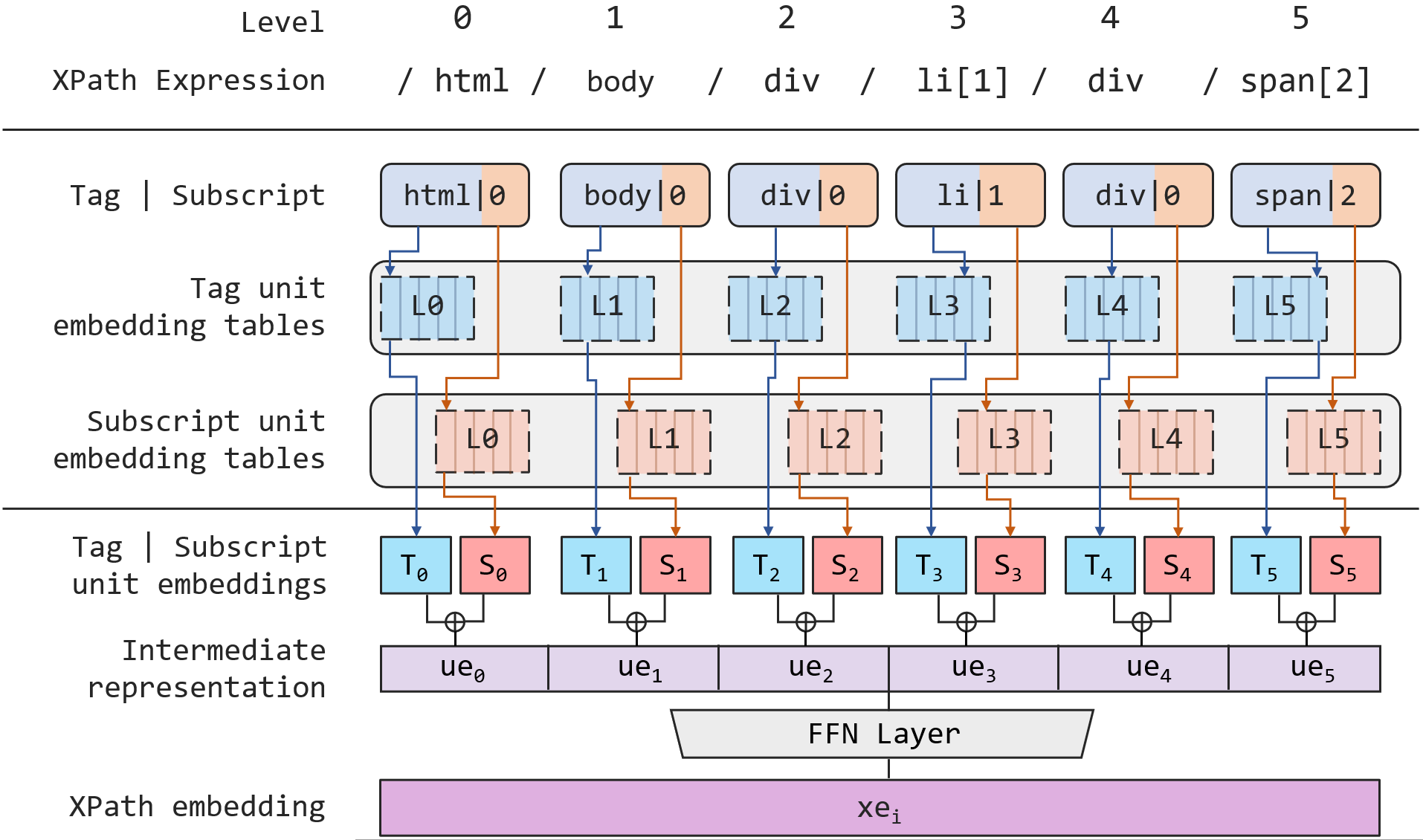}
    \caption{Overview of the XPath embedding from an XPath expression.}
    \label{fig:xpath_embedding}
\end{figure*}

To take advantage of existing pre-trained models and adapt to markup-language-based tasks (\textit{e.g.}, webpage tasks), we use the BERT~\citep{devlin-etal-2019-bert} architecture as the encoder backbone and add a new input embedding named \textbf{XPath embedding} to the original embedding layer. The overview structures of MarkupLM and the newly-proposed XPath Embedding are shown in Figure \ref{fig:markuplm} and \ref{fig:xpath_embedding}.

\paragraph{XPath Embedding}
For the $i$-th input token $x_i$, we take its corresponding XPath expression and split it by "/" to get the node information at each level of the XPath as a list, $xp_i=[(t_0^i,s^i_0),(t^i_1,s^i_1),\cdots,(t^i_d,s^i_d)]$, where $d$ is the depth of this XPath and $(t^i_j,s^i_j)$ denotes the tag name and the subscript of the XPath unit on level $j$ for $x_i$. Note that for units with no subscripts, we assign 0 to $s^i_j$. To facilitate further processing, we do truncation and padding on ${xp}_i$ to unify their lengths as $L$.

The process of converting XPath expression into XPath embedding is shown in Figure~\ref{fig:xpath_embedding}. For $(t^i_j,s^i_j)$, we input this pair into the $j$-th tag unit embedding table and $j$-th subscript unit embedding table respectively, and they are added up to get the $j$-th unit embedding $ue^i_j$. We set the dimensions of these two embeddings as $d_u$.
$$
ue^i_j = \texttt{TagUnitEmb}_j(t^i_j)+ \texttt{SubsUnitEmb}_j(s^i_j)
$$
We concatenate all the unit embeddings to get the intermediate representation $r_i$ of the complete XPath for $x_i$. 
$$
r_i = [ue^i_0;ue^i_1;\cdots;ue^i_{L}]
$$

Finally, to match the dimension of other embeddings, we feed the intermediate representation $r_i$ into an FFN layer to get the final XPath embedding $xe_i$.
\begin{gather*}
xe_i = W_2[\textrm{ReLU}(W_1r_i+b_1)]+b_2, \\
W_1\in \mathbb{R}^{4d_h\times Ld_u},b_1\in \mathbb{R}^{4d_h},\\
W_2\in \mathbb{R}^{d_h\times 4d_h},b_2\in \mathbb{R}^{d_h}
\end{gather*}
where $d_h$ is the hidden size of MarkupLM. To simplify the converting process, we have also tried replacing the FFN layer with a single linear transformation. However, this tiny modification makes the training process much more unstable and slightly hurts the performance so we keep the original design.

\subsection{Pre-training Objectives}
\label{section:pre-training}

To efficiently capture the complex structures of markup-language-based documents, we propose pre-training objectives on three different levels, including token-level (MMLM), node-level (NRP), and page-level (TPM).

\paragraph{Masked Markup Language Modeling}
Inspired by the previous works~\citep{devlin-etal-2019-bert, 10.1145/3394486.3403172, xu2021layoutlmv2}, we propose a token-level pre-training objective Masked Markup Language Modeling (MMLM), which is designed to enhance the language modeling ability with the markup clues. Basically, with the text and markup input sequences, we randomly select and replace some tokens with \texttt{[MASK]}, and this task requires the model to recover the masked tokens with all markup clues. 

\paragraph{Node Relation Prediction}
Although the MMLM task can help the model improve the markup language modeling ability, the model is still not aware of the semantics of XPath information provided by the XPath embedding. With the naturally structural DOM tree, we propose a node-level pre-training objective Node Relation Prediction (NRP) to explicitly model the relationship between a pair of nodes. We firstly define a set of directed node relationships $R$ $\in$ \{\texttt{self},  \texttt{parent}, \texttt{child}, \texttt{sibling}, \texttt{ancestor}, \texttt{descendent}, \texttt{others}\}. Then we combine each node to obtain the node pairs. For each pair of nodes, we assign the corresponding label according to the node relationship set, and the model is required to predict the assigned relationship labels with the features from the first token of each node.

\paragraph{Title-Page Matching}
Besides the fine-grained information provided by markups, the sentence-level or topic-level information can also be leveraged in markup-language-based documents. For HTML-based documents, the element \texttt{<title>} can be excellent summaries of the \texttt{<body>}, which provides a supervision for high-level semantics. To efficiently utilize this self-supervised information, we propose a page-level pre-training objective Title-Page Matching (TPM). Given the element \texttt{<body>} of a markup-based document, we randomly replace the text of element \texttt{<title>} and ask the model to predict if the title is replaced by using the representation of token \texttt{[CLS]} for binary classification.

\subsection{Fine-tuning}
We follow the scheme of common pre-trained language models~\citep{devlin-etal-2019-bert, Liu2019RoBERTaAR} and introduce the fine-tuning recipes on two downstream tasks including reading comprehension and information extraction.

For the reading comprehension task, we model it as an extractive QA task. The question and context are concatenated together as the input sequence, and slicing is required when its length exceeds a threshold. For tokens of questions, the corresponding XPath embeddings are the same as $\texttt{[PAD]}$ token. We input the last hidden state of each token to a binary linear classification layer to get two scores for start and end positions, and make span predictions with these scores following the common practice in SQuAD \cite{rajpurkar2016squad}. 

For the information extraction task, we model it as a token classification task. We input the last hidden state of each token to a linear classification layer, which has $n+1$ categories, where $n$ is the number of attributes we need to extract and the extra category is for tokens that belong to none of these attributes.

\section{Experiments}
In this work, we apply our MarkupLM framework to HTML-based webpages, which is one of the most common markup language scenarios. Equipped with the existing webpage datasets Common Crawl (CC)\footnote{\url{https://commoncrawl.org/}}, we pre-train MarkupLM with large-scale unlabeled HTML data and evaluate the pre-trained models on web-based structural reading comprehension and information extraction tasks.

% Table generated by Excel2LaTeX from sheet 'Sheet2'
\begin{table*}[ht]
%\small
  \centering
    \begin{tabular}{llccc}
     \toprule
    Model & Modality  &EM & F1    & POS \\
    \midrule
    T-PLM ($\textrm{BERT}_{\rm BASE}$) & Text  & 52.12 & 61.57 & 79.74 \\
    H-PLM ($\textrm{BERT}_{\rm BASE}$)  & Text + HTML  &61.51 & 67.04 & 82.97 \\
    V-PLM ($\textrm{BERT}_{\rm BASE}$) & Text + HTML + Image  & 62.07 & 66.66 & 83.64 \\
    \hdashline
    T-PLM ($\textrm{RoBERTa}_{\rm BASE}$) & Text & 52.32 & 63.19 & 80.93 \\
    H-PLM ($\textrm{RoBERTa}_{\rm BASE}$)  & Text + HTML  & 62.77 & 68.19 & 83.13 \\
    \midrule
    $\textrm{MarkupLM}_{\rm BASE}$ & Text + HTML  & \textbf{68.39} & \textbf{74.47} & \textbf{87.93} \\
    \midrule
    T-PLM ($\textrm{ELECTRA}_{\rm LARGE}$)  & Text  & 61.67 & 69.85 & 84.15 \\
    H-PLM ($\textrm{ELECTRA}_{\rm LARGE}$) & Text + HTML  & 70.12 & 74.14 & 86.33 \\
    V-PLM ($\textrm{ELECTRA}_{\rm LARGE}$) & Text + HTML + Image  & 73.22 & 76.16 & 87.06 \\
    \hdashline
    T-PLM ($\textrm{RoBERTa}_{\rm LARGE}$) & Text & 58.50 & 70.13 & 83.31 \\
    H-PLM ($\textrm{RoBERTa}_{\rm LARGE}$)  & Text + HTML  & 69.57 & 74.13 & 85.93 \\
    \midrule
    $\textrm{MarkupLM}_{\rm LARGE}$ & Text + HTML  & \textbf{74.43} &  \textbf{80.54} & \textbf{90.15} \\
    \bottomrule
    \end{tabular}%
  \caption{Evaluation results on the WebSRC development set. Results on BERT and ELECTRA are obtained from the original paper~\citep{chen-etal-2021-websrc}, while those on RoBERTa are our re-running.}
  \label{tab:websrc}%
\end{table*}%

\begin{table*}[ht]
   % \small
    \centering
    \begin{tabular}{cccccc}
    \toprule
        Model $\backslash$ \#Seed Sites & $k=1$ &  $k=2$ &  $k=3$ &  $k=4$ &  $k=5$\\
    \midrule
        \texttt{SSM} \citep{carlson2008bootstrapping} & 63.00 & 64.50 & 69.20 & 71.90 & 74.10\\
        \texttt{Render-Full} \citep{10.1145/2009916.2010020} & 84.30 & 86.00 & 86.80 & 88.40 & 88.60 \\
        \texttt{FreeDOM-NL} \citep{10.1145/3394486.3403153} & 72.52 & 81.33 & 86.44 & 88.55 & 90.28 \\
        \texttt{FreeDOM-Full} \citep{10.1145/3394486.3403153} & 82.32 & 86.36 & 90.49 & 91.29 & 92.56 \\
        \texttt{SimpDOM} \citep{zhou2021simplified} & 83.06 & 88.96 & 91.63 & 92.84 & 93.75\\
	\midrule
    	$\textrm{MarkupLM}_{\rm BASE}$ & 82.11& 91.29 & 94.42 & 95.31 & 95.89 \\
    	$\textrm{MarkupLM}_{\rm LARGE}$ & \textbf{85.71} & \textbf{93.57} & \textbf{96.12} & \textbf{96.71} & \textbf{97.37}\\
    \bottomrule 
    \end{tabular}
    \caption{Comparing the extraction performance (F1 score) of five baseline models to our method MarkupLM using different numbers of seed sites $k=\{1,2,3,4,5\}$ on the SWDE dataset, the results are from \citep{zhou2021simplified}. Each value in the table is computed from the average over 8 verticals and 10 permutations of seed websites per vertical (80 experiments in total).}
    \label{tab:freedom}
\end{table*}

\subsection{Data}
\paragraph{Common Crawl}
The Common Crawl (CC) dataset contains petabytes of webpages in the form of raw web page data, metadata extracts, and text extracts. We choose one of its snapshots\footnote{\url{https://commoncrawl.org/2021/08/july-august-2021-crawl-archive-available/}}, and use the pre-trained language detection model from \texttt{fasttext}~\citep{joulin2017bag} to filter out non-English pages. Specifically, we only take the page when the model predicts it as English with the classifier score > 0.6 and discard all the others. Besides, we only keep the tags that may contain texts (\textit{e.g.} \texttt{<div>}, \texttt{<span>}, \texttt{<li>}, \texttt{<a>}, etc.) and delete those with no texts (\textit{e.g.,} \texttt{<script>}, \texttt{<style>}, etc.) in these pages to save storage space. After pre-processing, a subset of CC with 24M English webpages is extracted as our pre-training data for MarkupLM.

\paragraph{WebSRC} 
The Web-based Structural Reading Comprehension (WebSRC) dataset~\citep{chen-etal-2021-websrc} consists of 440K question-answer pairs, which are collected from 6.5K web pages with corresponding HTML source code, screenshots, and metadata. Each question in WebSRC requires a certain structural understanding of a webpage to answer, and the answer is either a text span on the web page or yes/no. After adding the additional yes/no tokens to the text input, WebSRC can be modeled as a typical extractive reading comprehension task. Following the original paper~\citep{chen-etal-2021-websrc}, we choose evaluation metrics for this dataset as \textbf{Exact match (EM)}, \textbf{F1 score (F1)}, and \textbf{Path overlap score (POS)}. We use the official split to get the training and development set. Note that the authors of WebSRC did not release their testing set, so all our results are obtained from the development set.

\paragraph{SWDE}
The Structured Web Data Extraction (SWDE) dataset~\citep{10.1145/2009916.2010020} is a real-world webpage collection for automatic extraction of structured data from the Web. It involves 8 verticals, 80 websites (10 per vertical), and 124,291 webpages (200 - 2,000 per website) in total. The task is to extract the values corresponding to a set of given attributes (depending on which vertical the webpage belongs to) from a webpage, like value for \textit{author} in \textit{book} pages. Following previous works~\citep{10.1145/2009916.2010020, 10.1145/3394486.3403153, zhou2021simplified}, we choose \textbf{page-level F1 scores} as our evaluation metrics for this dataset.

Since there is no official train-test split, we follow previous works~\citep{10.1145/2009916.2010020, 10.1145/3394486.3403153, zhou2021simplified} to do training and evaluation on each vertical ($i.e.$, category of websites) independently. In each vertical, we select $k$ consecutive seed websites as the training data and use the remaining $10-k$ websites as the testing set. Note that in this few-shot extraction task, none of the pages in the $10-k$ websites have been visited in the training phase. This setting is abstracted from the real application scenario where only a small set of labeled data is provided for specific websites and we aim to infer the attributes on a much larger unseen website set. The final results are obtained by taking the average of all 8 verticals and all 10 permutations of seed websites per vertical, leading to 80 individual experiments for each $k$. For the pre- and post-processing of data, we follow \citet{zhou2021simplified} to make a fair comparison.

\begin{table*}[t]
\begin{minipage}{0.5\linewidth}

  \centering\small
  \setlength{\tabcolsep}{6pt}

    \begin{tabular}{cccccc}
    \toprule
    Ver. $\backslash$ \#Seed & $k=1$  & $k=2$   & $k=3$   & $k=4$   & $k=5$ \\
    \midrule
    auto    & 70.63 & 89.08      & 94.73      & 95.45      &    98.15     \\
    book   & 81.89  & 87.43 & 89.40      & 90.26      & 90.35       \\
    camera    & 84.65 & 92.72 & 94.63 & 95.16      &    94.99     \\
    job    & 76.86 & 86.19 & 90.02 & 90.99      &    92.34     \\
    movie & 90.53   & 94.87 & 97.85 & 98.91 & 99.37             \\
    nbaplayer & 85.92   & 91.97 & 94.31 & 94.15 & 96.07          \\
    restaurant & 82.76   & 92.25  & 95.87 & 98.70 &  97.04         \\
    university & 83.67   & 95.80 & 98.55 & 98.82 & 98.77         \\
    \midrule
    \textbf{Average} & 82.11 & 91.29 & 94.42 & 95.31 & 95.89  \\

    \bottomrule
    \end{tabular}
    \end{minipage}
    \begin{minipage}{0.5\linewidth}
  \centering\small
  \setlength{\tabcolsep}{5pt}

    \begin{tabular}{cccccc}
    \toprule
    Ver. $\backslash$ \#Seed & $k=1$  & $k=2$   & $k=3$   & $k=4$   & $k=5$ \\
    \midrule
    auto    & 74.77 & 86.88      & 96.22      & 96.46      &    99.19     \\
    book   & 85.73  & 92.01 & 92.97      & 93.29      & 93.46        \\
    camera    & 85.18 & 95.09 & 96.22 & 96.69      &    96.27     \\
    job    & 80.64 & 90.67 & 90.41 & 90.72      &    92.99     \\
    movie & 94.27   & 98.55 & 99.23 & 99.66 & 99.58             \\
    nbaplayer & 88.95   & 94.27 & 97.76 & 98.26 & 98.77          \\
    restaurant & 87.06   & 94.37  & 98.06 & 98.7 &  98.83         \\
    university & 89.10   & 96.69 & 98.07 & 99.87 & 99.88         \\
    \midrule
    \textbf{Average} & 85.71 & 93.57 & 96.12 & 96.71 & 97.37  \\

    \bottomrule
    \end{tabular}

    \end{minipage}
    
\caption{Evaluation results of $\textrm{MarkupLM}$ ($\textrm{BASE}$ on left and $\textrm{LARGE}$ on right) on the SWDE dataset with different numbers of seed sites $k=\{1,2,3,4,5\}$ for training, Ver. stands for vertical while \#Seed is the number of seed sites.}
  \label{tab:freedom_detail}%
\end{table*}%

\subsection{Settings}
\label{sec:settings}
%\subsubsection{Hyperparameters}
\paragraph{Pre-training}
The size of the selected tags and subscripts in XPath embedding are 216 and 1,001 respectively, the max depth of XPath expression ($L$) is 50, and the dimension for the tag-unit and subscript-unit embedding ($d_u$) is 32. The token-masked probability in MMLM and title-replaced probability in TPM are both 15\%, and we do not mask the tokens in the input sequence corresponding to the webpage titles. The max number of selected node pairs is 1,000 in NRP for each sample, and we limit the ratio of pairs with \texttt{non-others} (\textit{i.e.}, \texttt{self, parent, $\cdots$}) labels as 80\% to make a balance. We initialize $\textrm{MarkupLM}$ from $\textrm{RoBERTa}$ and train it for 300K steps on 8 NVIDIA A100 GPUs. We set the total batch size as 256, the learning rate as 5e-5, and the warmup ratio as 0.06. The selected optimizer is AdamW \cite{loshchilov2018decoupled}, with $\epsilon=1e-6$, $\beta_1=0.9$, $\beta_2=0.98$, $\texttt{weight decay}=0.01$, and a linear decay learning rate scheduler with 6\% warmup steps. We also apply $\texttt{FP16}$, $\texttt{gradient-checkpointing}$~\citep{chen2016training}, and $\texttt{deepspeed}$~\citep{rasley2020deepspeed} to reduce GPU memory consumption and accelerate training.

\paragraph{Fine-tuning}
For WebSRC, we fine-tune MarkupLM for 5 epochs with the total batch size of 64, the learning rate of 1e-5, and the warmup ratio of 0.1. For SWDE, we fine-tune MarkupLM with 10 epochs, the total batch size of 64, the learning rate of 2e-5, and the warmup ratio of 0.1. The max sequence length is set as 384 in both tasks, and we keep other hyper-parameters as default.

\subsection{Results}
The results for WebSRC are shown in Table \ref{tab:websrc}. Selected baselines are T-PLM, H-PLM, and V-PLM in \citet{chen-etal-2021-websrc}, referring to the paper for more details. To make a fair comparison, we re-run the released baseline experiments with RoBERTa. We observe $\textrm{MarkupLM}$ significantly surpass H-PLM which uses the same modality of information. This strongly indicates that MarkupLM makes better use of the XPath features with the specially designed embedding layer and pre-training objectives compared with merely adding more tag tokens into the input sequence as in H-PLM. Besides, $\textrm{MarkupLM}$ also achieves a higher score than the previous state-of-the-art V-PLM model that requires a huge amount of external resources to render the HTML source codes and uses additional vision features from Faster R-CNN~\citep{NIPS2015_14bfa6bb}, showing that our render-free MarkupLM is more lightweight and can learn the structural information better even without any visual information. It is also worth noting that adding HTML tags as input tokens in H-PLM and V-PLM drastically increases the length of input strings, so more slicing operations are required to fit the length limitation of language models, which results in more training samples ($\sim$860k) and longer training time, while MarkupLM does not suffer from this (only $\sim$470k training samples) and can greatly reduce training time.

The results for SWDE are in Table \ref{tab:freedom} and \ref{tab:freedom_detail}. It is observed that our $\textrm{MarkupLM}$ also substantially outperforms the strong baselines. Different from the previous state-of-the-art model SimpDOM which explicitly sends the relationship between DOM tree nodes into their model and adds huge amounts of extra discrete features (\textit{e.g.}, whether a node contains numbers or dates), MarkupLM is much simpler and is free from time-consuming additional webpage annotations. We also report detailed statistics with regard to different verticals in Table \ref{tab:freedom_detail}. With the growth of $k$, MarkupLM gets more webpages as the training set, so there is a clear ascending trend reflected by the scores. We also see the variance among different verticals since the number and type of pages are not the same.

% Table generated by Excel2LaTeX from sheet 'ablation'

%\begin{table*}[htbp]
%  \centering

%    \begin{tabular}{cccccccc}
%    \toprule
%    Initialization & Data  & MMLM   & NRP   & TPM   & Exact Match & F1    & POS \\
%    \midrule
%    $\textrm{BERT}_{\rm BASE}$ & 1M    & \checkmark &       &       &  54.11     &  63.44     & 81.87 \\
%    $\textrm{BERT}_{\rm BASE}$ & 1M    & \checkmark & \checkmark &       &   56.72    & 65.07      & 83.02 \\
%    $\textrm{BERT}_{\rm BASE}$ & 1M    & \checkmark & \checkmark & \checkmark &  59.56     &  68.12     &  84.80\\
%    $\textrm{RoBERTa}_{\rm BASE}$ & 1M    & \checkmark & \checkmark & \checkmark &  60.83     &  69.13     & 85.61 \\
%    $\textrm{RoBERTa}_{\rm BASE}$ & 24M   & \checkmark & \checkmark & \checkmark &  67.38     &   74.80    & 87.24 \\
%    \bottomrule
%    \end{tabular}%
%    \caption{Ablation study with different pre-trained objectives and initialization for $\textrm{MarkupLM}_{\rm BASE}$ on the WebSRC dataset.}
%  \label{tab:ablation}%
%\end{table*}%

% Table generated by Excel2LaTeX from sheet 'ablationv3'
\begin{table*}[ht]
  \centering
  
    \begin{tabular}{ccccccccc}
    \toprule
          &  & Pre-training Data & \multicolumn{3}{c}{Objectives} & \multicolumn{3}{c}{Metrics} \\
          \midrule
          \# & Initialization & Samples   & MMLM   & NRP   & TPM   & EM & F1    & POS \\
    \midrule
    %1     & $\textrm{BERT}_{\rm BASE}$      & 0      & Text      &       &       & & 49.70   & 59.81 & 77.65 \\
    %\midrule
    %2     & $\textrm{BERT}_{\rm BASE}$ & 1M    & Text  & \checkmark &       &       & 49.62 & 59.95  & 78.54 \\
    %\midrule
    1a     & $\textrm{BERT}_{\rm BASE}$ & 1M     & \checkmark & & &  54.29 & 61.47 & 82.03  \\
    1b     & $\textrm{BERT}_{\rm BASE}$ & 1M     & \checkmark & \checkmark      &  & 56.72       & 65.07 & 83.02 \\
    1c     & $\textrm{BERT}_{\rm BASE}$ & 1M     & \checkmark &  &  \checkmark     &  58.87     &  66.74     & 83.85 \\
    1d     & $\textrm{BERT}_{\rm BASE}$ & 1M     & \checkmark & \checkmark & \checkmark & 59.56      &  68.12     & 84.80 \\
    \midrule
    2a     & $\textrm{RoBERTa}_{\rm BASE}$ & 1M     & \checkmark & \checkmark & \checkmark & 61.48      & 69.15      & 84.32 \\
    2b     & $\textrm{RoBERTa}_{\rm BASE}$ & 24M   & \checkmark & \checkmark & \checkmark & 68.39 & 74.47 & 87.93  \\
    \bottomrule
    \end{tabular}%
    \caption{Ablation study on the WebSRC dataset, where EM, F1 and POS scores on the development set are reported. "MMLM", "NRP" and "TPM“ stand for Masked Markup Language Model, Node Relation Prediction and Title Page Matching respectively. All these models, except \#2b, are pre-trained with 200k steps and the same hyper-parameter settings described in Section \ref{sec:settings}.}
  \label{tab:ablation}%
\end{table*}%

\subsection{Ablation Study}
To investigate how each pre-training objective contributes to MarkupLM, we conduct an ablation study on WebSRC with a smaller training set containing 1M webpages. The model we initialized from is BERT-base-uncased in this sub-section with all the other settings unchanged. The results are in Table \ref{tab:ablation}. According to the four results in \#1, we see both of the newly-proposed training objectives improve the model performance substantially, and the proposed TPM (+4.6\%EM) benefits the model more than NRP (+2.4\%EM). Using both objectives together is more effective than using either one alone, leading to an increase of 5.3\% on EM. We can also see a performance improvement (+1.9\%EM) from \#1d to \#2a when replacing BERT with a stronger initial model RoBERTa. Finally, we get the best model with all three objectives and better initialization on larger data, as the comparison between \#2a and \#2b.

%the NRP objective that models the relationship between nodes in the DOM tree greatly helps the MarkupLM to learn the structural information. Furthermore, after adding the page-level TPM objective (lines 2 and 3), the performance is enhanced to a higher level. We hold that this objective is able to help MarkupLM learn the semantic information of questions in this dataset. We also investigate the impact of different initialization by replacing BERT with RoBERTa and keeping all three objectives (lines 3 and 4) and confirm the benefits of initializing from a better model.

%Table \ref{tab:websrc} Table \ref{tab:freedom} Table \ref{tab:freedom_detail} Table \ref{tab:ablation}

\section{Related Work}

Multimodal pre-training with text, layout, and image information has significantly advanced the research of document AI, and it has been the de facto approach in a variety of VRDU tasks. Although great progress has been achieved for the fixed-layout document understanding tasks, the existing multimodal pre-training approaches cannot be easily applied to markup-based document understanding in a straightforward way, because the layout information of markup-based documents needs to be rendered dynamically and may be different depending on software and hardware. Therefore, the markup information is vital for the document understanding. \citet{ashby-weir-2020-leveraging} compared the Text+Tags approach with their Text-Only equivalents over five web-based NER datasets, which indicates the necessity of markup enrichment of deep language models. \citet{10.1145/3394486.3403153} presented a novel two-stage neural approach named FreeDOM. The first stage learns a representation for each DOM node in the page by combining both the text and markup information. The second stage captures longer range distance and semantic relatedness using a relational neural network. Experiments show that FreeDOM beats the previous SOTA results without requiring features over rendered pages or expensive hand-crafted features. \citet{zhou2021simplified} proposed a novel transferable method SimpDOM to tackle the problem by efficiently retrieving useful context for each node by leveraging the tree structure. \citet{xie2021webke} introduced a framework called WebKE that extracts knowledge triples from semi-structured webpages by extending pre-trained language models to markup language and encoding layout semantics.

However, these methods did not fully leverage the large-scale unlabeled data and self-supervised pre-training techniques to enrich the document representation learning. To the best of our knowledge, MarkupLM is the first large-scale pre-trained model that jointly learns the text and markup language in a single framework for VRDU tasks.

\section{Conclusion and Future Work}

In this paper, we present MarkupLM, a simple yet effective pre-training approach for text and markup language. With the Transformer architecture, MarkupLM integrates different input embeddings including text embeddings, positional embeddings, and XPath embeddings. Furthermore, we also propose new pre-training objectives that are specially designed for understanding the markup language. We evaluate the pre-trained MarkupLM model on the WebSRC and SWDE datasets. Experiments show that MarkupLM significantly outperforms several SOTA baselines in these tasks.

For future research, we will investigate the MarkupLM pre-training with more data and more computation resources, as well as the language expansion. Furthermore, we will also pre-train MarkupLM models for digital-born PDFs and Office documents that use XML DOM as the backbones. In addition, we will also explore the relationship between MarkupLM and layout-based models (like LayoutLM) to deeply understand whether these two kinds of models can be pre-trained under a unified multi-view and multi-task setting and whether the knowledge from these two kinds of models can be transferred to each other to better understand the structural information.

% Entries for the entire Anthology, followed by custom entries
\bibliography{anthology,custom}

\begin{thebibliography}{34}
\expandafter\ifx\csname natexlab\endcsname\relax\def\natexlab#1{#1}\fi

\bibitem[{Appalaraju et~al.(2021)Appalaraju, Jasani, Kota, Xie, and
  Manmatha}]{appalaraju2021docformer}
Srikar Appalaraju, Bhavan Jasani, Bhargava~Urala Kota, Yusheng Xie, and
  R.~Manmatha. 2021.
\newblock \href {http://arxiv.org/abs/2106.11539} {Docformer: End-to-end
  transformer for document understanding}.

\bibitem[{Ashby and Weir(2020)}]{ashby-weir-2020-leveraging}
Colin Ashby and David Weir. 2020.
\newblock \href {https://doi.org/10.18653/v1/2020.coling-main.36} {Leveraging
  {HTML} in free text web named entity recognition}.
\newblock In \emph{Proceedings of the 28th International Conference on
  Computational Linguistics}, pages 407--413, Barcelona, Spain (Online).
  International Committee on Computational Linguistics.

\bibitem[{Carlson and Schafer(2008)}]{carlson2008bootstrapping}
Andrew Carlson and Charles Schafer. 2008.
\newblock Bootstrapping information extraction from semi-structured web pages.
\newblock In \emph{Joint European Conference on Machine Learning and Knowledge
  Discovery in Databases}, pages 195--210. Springer.

\bibitem[{Chen et~al.(2016)Chen, Xu, Zhang, and Guestrin}]{chen2016training}
Tianqi Chen, Bing Xu, Chiyuan Zhang, and Carlos Guestrin. 2016.
\newblock Training deep nets with sublinear memory cost.
\newblock \emph{arXiv preprint arXiv:1604.06174}.

\bibitem[{Chen et~al.(2021)Chen, Zhao, Chen, Ji, Zhang, Luo, Xiong, and
  Yu}]{chen-etal-2021-websrc}
Xingyu Chen, Zihan Zhao, Lu~Chen, JiaBao Ji, Danyang Zhang, Ao~Luo, Yuxuan
  Xiong, and Kai Yu. 2021.
\newblock \href {https://aclanthology.org/2021.emnlp-main.343} {{W}eb{SRC}: A
  dataset for web-based structural reading comprehension}.
\newblock In \emph{Proceedings of the 2021 Conference on Empirical Methods in
  Natural Language Processing}, pages 4173--4185, Online and Punta Cana,
  Dominican Republic. Association for Computational Linguistics.

\bibitem[{Devlin et~al.(2019)Devlin, Chang, Lee, and
  Toutanova}]{devlin-etal-2019-bert}
Jacob Devlin, Ming-Wei Chang, Kenton Lee, and Kristina Toutanova. 2019.
\newblock \href {https://doi.org/10.18653/v1/N19-1423} {{BERT}: Pre-training of
  deep bidirectional transformers for language understanding}.
\newblock In \emph{Proceedings of the 2019 Conference of the North {A}merican
  Chapter of the Association for Computational Linguistics: Human Language
  Technologies, Volume 1 (Long and Short Papers)}, pages 4171--4186,
  Minneapolis, Minnesota. Association for Computational Linguistics.

\bibitem[{Graliński et~al.(2020)Graliński, Stanisławek, Wróblewska,
  Lipiński, Kaliska, Rosalska, Topolski, and Biecek}]{graliski2020kleister}
Filip Graliński, Tomasz Stanisławek, Anna Wróblewska, Dawid Lipiński,
  Agnieszka Kaliska, Paulina Rosalska, Bartosz Topolski, and Przemysław
  Biecek. 2020.
\newblock \href {http://arxiv.org/abs/2003.02356} {Kleister: A novel task for
  information extraction involving long documents with complex layout}.

\bibitem[{Hao et~al.(2011)Hao, Cai, Pang, and Zhang}]{10.1145/2009916.2010020}
Qiang Hao, Rui Cai, Yanwei Pang, and Lei Zhang. 2011.
\newblock \href {https://doi.org/10.1145/2009916.2010020} {From one tree to a
  forest: a unified solution for structured web data extraction}.
\newblock In \emph{Proceeding of the 34th International {ACM} {SIGIR}
  Conference on Research and Development in Information Retrieval, {SIGIR}
  2011, Beijing, China, July 25-29, 2011}, pages 775--784. {ACM}.

\bibitem[{Harley et~al.(2015)Harley, Ufkes, and Derpanis}]{harley2015icdar}
Adam~W Harley, Alex Ufkes, and Konstantinos~G Derpanis. 2015.
\newblock Evaluation of deep convolutional nets for document image
  classification and retrieval.
\newblock In \emph{International Conference on Document Analysis and
  Recognition ({ICDAR})}.

\bibitem[{Hong et~al.(2021)Hong, Kim, Ji, Hwang, Nam, and Park}]{hong2021bros}
Teakgyu Hong, DongHyun Kim, Mingi Ji, Wonseok Hwang, Daehyun Nam, and Sungrae
  Park. 2021.
\newblock \href {https://openreview.net/forum?id=punMXQEsPr0} {{BROS}: A
  pre-trained language model for understanding texts in document}.

\bibitem[{{Huang} et~al.(2019){Huang}, {Chen}, {He}, {Bai}, {Karatzas}, {Lu},
  and {Jawahar}}]{8977955}
Z.~{Huang}, K.~{Chen}, J.~{He}, X.~{Bai}, D.~{Karatzas}, S.~{Lu}, and C.~V.
  {Jawahar}. 2019.
\newblock \href {https://doi.org/10.1109/ICDAR.2019.00244} {Icdar2019
  competition on scanned receipt ocr and information extraction}.
\newblock In \emph{2019 International Conference on Document Analysis and
  Recognition (ICDAR)}, pages 1516--1520.

\bibitem[{Jaume et~al.(2019)Jaume, Ekenel, and Thiran}]{Jaume2019FUNSDAD}
Guillaume Jaume, Hazim~Kemal Ekenel, and Jean-Philippe Thiran. 2019.
\newblock Funsd: A dataset for form understanding in noisy scanned documents.
\newblock \emph{2019 International Conference on Document Analysis and
  Recognition Workshops (ICDARW)}, 2:1--6.

\bibitem[{Joulin et~al.(2017)Joulin, Grave, Bojanowski, and
  Mikolov}]{joulin2017bag}
Armand Joulin, Edouard Grave, Piotr Bojanowski, and Tomas Mikolov. 2017.
\newblock \href {https://aclanthology.org/E17-2068} {Bag of tricks for
  efficient text classification}.
\newblock In \emph{Proceedings of the 15th Conference of the {E}uropean Chapter
  of the Association for Computational Linguistics: Volume 2, Short Papers},
  pages 427--431, Valencia, Spain. Association for Computational Linguistics.

\bibitem[{Lewis et~al.(2006)Lewis, Agam, Argamon, Frieder, Grossman, and
  Heard}]{10.1145/1148170.1148307}
D.~Lewis, G.~Agam, S.~Argamon, O.~Frieder, D.~Grossman, and J.~Heard. 2006.
\newblock \href {https://doi.org/10.1145/1148170.1148307} {Building a test
  collection for complex document information processing}.
\newblock In \emph{Proceedings of the 29th Annual International ACM SIGIR
  Conference on Research and Development in Information Retrieval}, SIGIR '06,
  page 665–666, New York, NY, USA. Association for Computing Machinery.

\bibitem[{Li et~al.(2021{\natexlab{a}})Li, Bi, Yan, Wang, Huang, Huang, and
  Si}]{li2021structurallm}
Chenliang Li, Bin Bi, Ming Yan, Wei Wang, Songfang Huang, Fei Huang, and Luo
  Si. 2021{\natexlab{a}}.
\newblock \href {https://doi.org/10.18653/v1/2021.acl-long.493}
  {{S}tructural{LM}: Structural pre-training for form understanding}.
\newblock In \emph{Proceedings of the 59th Annual Meeting of the Association
  for Computational Linguistics and the 11th International Joint Conference on
  Natural Language Processing (Volume 1: Long Papers)}, pages 6309--6318,
  Online. Association for Computational Linguistics.

\bibitem[{Li et~al.(2021{\natexlab{b}})Li, Gu, Kuen, Morariu, Zhao, Jain,
  Manjunatha, and Liu}]{li2021selfdoc}
Peizhao Li, Jiuxiang Gu, Jason Kuen, Vlad~I. Morariu, Handong Zhao, Rajiv Jain,
  Varun Manjunatha, and Hongfu Liu. 2021{\natexlab{b}}.
\newblock \href {http://arxiv.org/abs/2106.03331} {Selfdoc: Self-supervised
  document representation learning}.

\bibitem[{Lin et~al.(2020)Lin, Sheng, Vo, and Tata}]{10.1145/3394486.3403153}
Bill~Yuchen Lin, Ying Sheng, Nguyen Vo, and Sandeep Tata. 2020.
\newblock \href {https://dl.acm.org/doi/10.1145/3394486.3403153} {Freedom: {A}
  transferable neural architecture for structured information extraction on web
  documents}.
\newblock In \emph{{KDD} '20: The 26th {ACM} {SIGKDD} Conference on Knowledge
  Discovery and Data Mining, Virtual Event, CA, USA, August 23-27, 2020}, pages
  1092--1102. {ACM}.

\bibitem[{Liu et~al.(2019)Liu, Ott, Goyal, Du, Joshi, Chen, Levy, Lewis,
  Zettlemoyer, and Stoyanov}]{Liu2019RoBERTaAR}
Yinhan Liu, Myle Ott, Naman Goyal, Jingfei Du, Mandar Joshi, Danqi Chen, Omer
  Levy, Mike Lewis, Luke~S. Zettlemoyer, and Veselin Stoyanov. 2019.
\newblock \href {https://arxiv.org/abs/1907.11692} {Roberta: A robustly
  optimized bert pretraining approach}.
\newblock \emph{ArXiv preprint}, abs/1907.11692.

\bibitem[{Loshchilov and Hutter(2019)}]{loshchilov2018decoupled}
Ilya Loshchilov and Frank Hutter. 2019.
\newblock \href {https://openreview.net/forum?id=Bkg6RiCqY7} {Decoupled weight
  decay regularization}.
\newblock In \emph{7th International Conference on Learning Representations,
  {ICLR} 2019, New Orleans, LA, USA, May 6-9, 2019}. OpenReview.net.

\bibitem[{Mathew et~al.(2020)Mathew, Karatzas, Manmatha, and
  Jawahar}]{mathew2020docvqa}
Minesh Mathew, Dimosthenis Karatzas, R.~Manmatha, and C.~V. Jawahar. 2020.
\newblock \href {http://arxiv.org/abs/2007.00398} {Docvqa: A dataset for vqa on
  document images}.

\bibitem[{Park et~al.(2019)Park, Shin, Lee, Lee, Surh, Seo, and
  Lee}]{park2019cord}
Seunghyun Park, Seung Shin, Bado Lee, Junyeop Lee, Jaeheung Surh, Minjoon Seo,
  and Hwalsuk Lee. 2019.
\newblock \href {https://openreview.net/forum?id=SJl3z659UH} {{CORD}: A
  consolidated receipt dataset for post-{OCR} parsing}.
\newblock In \emph{Workshop on Document Intelligence at NeurIPS 2019}.

\bibitem[{Powalski et~al.(2021)Powalski, Łukasz Borchmann, Jurkiewicz, Dwojak,
  Pietruszka, and Pałka}]{powalski2021going}
Rafał Powalski, Łukasz Borchmann, Dawid Jurkiewicz, Tomasz Dwojak, Michał
  Pietruszka, and Gabriela Pałka. 2021.
\newblock \href {http://arxiv.org/abs/2102.09550} {Going full-tilt boogie on
  document understanding with text-image-layout transformer}.

\bibitem[{Pramanik et~al.(2020)Pramanik, Mujumdar, and
  Patel}]{pramanik2020multimodal}
Subhojeet Pramanik, Shashank Mujumdar, and Hima Patel. 2020.
\newblock \href {http://arxiv.org/abs/2009.14457} {Towards a multi-modal,
  multi-task learning based pre-training framework for document representation
  learning}.

\bibitem[{Rajpurkar et~al.(2016)Rajpurkar, Zhang, Lopyrev, and
  Liang}]{rajpurkar2016squad}
Pranav Rajpurkar, Jian Zhang, Konstantin Lopyrev, and Percy Liang. 2016.
\newblock Squad: 100,000+ questions for machine comprehension of text.
\newblock In \emph{Proceedings of the 2016 Conference on Empirical Methods in
  Natural Language Processing}, pages 2383--2392.

\bibitem[{Rasley et~al.(2020)Rasley, Rajbhandari, Ruwase, and
  He}]{rasley2020deepspeed}
Jeff Rasley, Samyam Rajbhandari, Olatunji Ruwase, and Yuxiong He. 2020.
\newblock Deepspeed: System optimizations enable training deep learning models
  with over 100 billion parameters.
\newblock In \emph{Proceedings of the 26th ACM SIGKDD International Conference
  on Knowledge Discovery \& Data Mining}, pages 3505--3506.

\bibitem[{Ren et~al.(2015)Ren, He, Girshick, and Sun}]{NIPS2015_14bfa6bb}
Shaoqing Ren, Kaiming He, Ross Girshick, and Jian Sun. 2015.
\newblock \href
  {https://proceedings.neurips.cc/paper/2015/file/14bfa6bb14875e45bba028a21ed38046-Paper.pdf}
  {Faster r-cnn: Towards real-time object detection with region proposal
  networks}.
\newblock In \emph{Advances in Neural Information Processing Systems},
  volume~28. Curran Associates, Inc.

\bibitem[{Vaswani et~al.(2017)Vaswani, Shazeer, Parmar, Uszkoreit, Jones,
  Gomez, Kaiser, and Polosukhin}]{vaswani2017attention}
Ashish Vaswani, Noam Shazeer, Niki Parmar, Jakob Uszkoreit, Llion Jones,
  Aidan~N. Gomez, Lukasz Kaiser, and Illia Polosukhin. 2017.
\newblock \href
  {https://proceedings.neurips.cc/paper/2017/hash/3f5ee243547dee91fbd053c1c4a845aa-Abstract.html}
  {Attention is all you need}.
\newblock In \emph{Advances in Neural Information Processing Systems 30: Annual
  Conference on Neural Information Processing Systems 2017, December 4-9, 2017,
  Long Beach, CA, {USA}}, pages 5998--6008.

\bibitem[{Wu et~al.(2021)Wu, Li, Zhang, Chen, Hombaiah, and
  Bendersky}]{wu2021lampret}
Te-Lin Wu, Cheng Li, Mingyang Zhang, Tao Chen, Spurthi~Amba Hombaiah, and
  Michael Bendersky. 2021.
\newblock \href {http://arxiv.org/abs/2104.08405} {Lampret: Layout-aware
  multimodal pretraining for document understanding}.

\bibitem[{Xie et~al.(2021)Xie, Huang, Liang, Huang, and Xiao}]{xie2021webke}
Chenhao Xie, Wenhao Huang, Jiaqing Liang, Chengsong Huang, and Yanghua Xiao.
  2021.
\newblock Webke: Knowledge extraction from semi-structured web with pre-trained
  markup language model.
\newblock In \emph{Proceedings of the 30th ACM International Conference on
  Information \& Knowledge Management}, pages 2211--2220.

\bibitem[{Xu et~al.(2021{\natexlab{a}})Xu, Xu, Lv, Cui, Wei, Wang, Lu,
  Florencio, Zhang, Che, Zhang, and Zhou}]{xu2021layoutlmv2}
Yang Xu, Yiheng Xu, Tengchao Lv, Lei Cui, Furu Wei, Guoxin Wang, Yijuan Lu,
  Dinei Florencio, Cha Zhang, Wanxiang Che, Min Zhang, and Lidong Zhou.
  2021{\natexlab{a}}.
\newblock \href {https://doi.org/10.18653/v1/2021.acl-long.201}
  {{L}ayout{LM}v2: Multi-modal pre-training for visually-rich document
  understanding}.
\newblock In \emph{Proceedings of the 59th Annual Meeting of the Association
  for Computational Linguistics and the 11th International Joint Conference on
  Natural Language Processing (Volume 1: Long Papers)}, pages 2579--2591,
  Online. Association for Computational Linguistics.

\bibitem[{Xu et~al.(2020)Xu, Li, Cui, Huang, Wei, and
  Zhou}]{10.1145/3394486.3403172}
Yiheng Xu, Minghao Li, Lei Cui, Shaohan Huang, Furu Wei, and Ming Zhou. 2020.
\newblock \href {https://dl.acm.org/doi/10.1145/3394486.3403172} {Layoutlm:
  Pre-training of text and layout for document image understanding}.
\newblock In \emph{{KDD} '20: The 26th {ACM} {SIGKDD} Conference on Knowledge
  Discovery and Data Mining, Virtual Event, CA, USA, August 23-27, 2020}, pages
  1192--1200. {ACM}.

\bibitem[{Xu et~al.(2021{\natexlab{b}})Xu, Lv, Cui, Wang, Lu, Florencio, Zhang,
  and Wei}]{xu2021layoutxlm}
Yiheng Xu, Tengchao Lv, Lei Cui, Guoxin Wang, Yijuan Lu, Dinei Florencio, Cha
  Zhang, and Furu Wei. 2021{\natexlab{b}}.
\newblock \href {http://arxiv.org/abs/2104.08836} {Layoutxlm: Multimodal
  pre-training for multilingual visually-rich document understanding}.

\bibitem[{Zhou et~al.(2021)Zhou, Sheng, Vo, Edmonds, and
  Tata}]{zhou2021simplified}
Yichao Zhou, Ying Sheng, Nguyen Vo, Nick Edmonds, and Sandeep Tata. 2021.
\newblock \href {http://arxiv.org/abs/2101.02415} {Simplified dom trees for
  transferable attribute extraction from the web}.

\bibitem[{Łukasz Garncarek et~al.(2021)Łukasz Garncarek, Powalski,
  Stanisławek, Topolski, Halama, Turski, and
  Graliński}]{garncarek2021lambert}
Łukasz Garncarek, Rafał Powalski, Tomasz Stanisławek, Bartosz Topolski,
  Piotr Halama, Michał Turski, and Filip Graliński. 2021.
\newblock \href {http://arxiv.org/abs/2002.08087} {Lambert: Layout-aware
  (language) modeling for information extraction}.

\end{thebibliography}
\bibliographystyle{acl_natbib}

% \appendix

% \section{Example Appendix}
% \label{sec:appendix}

%This is an appendix.

\end{document}